\def\year{2019}\relax
\documentclass[letterpaper]{article} 
\usepackage{aaai19}  
\usepackage{amsmath}
\usepackage{amsfonts}
\usepackage{amssymb}
\usepackage{times}  
\usepackage{helvet}  
\usepackage{courier}  
\usepackage{url}  
\usepackage{csquotes}
\usepackage{tabularx}
\usepackage{xcolor}
\usepackage[inline]{enumitem}

\usepackage{listings}
\usepackage{graphicx}  
\author{
Arindam Mitra$^\ddagger$ \and  Peter Clark$^\dagger$ \and Oyvind Tafjord$^\dagger$  \and  Chitta Baral$^\ddagger$\\
\\
$^\dagger$ Allen Institute for Artificial Intelligence, Seattle, WA, U.S.A.\\
$^\ddagger$ Arizona State University, AZ, U.S.A.\\
{\tt \small \{peterc,oyvindt\}@allenai.org,\{amitra7,chitta\}@asu.edu}
}
\begin{document}
%
\title{Declarative Question Answering over Knowledge Bases containing Natural Language Text with Answer Set Programming}

\maketitle
\begin{abstract}
While in recent years machine learning (ML) based approaches have been the popular approach in developing end-to-end question answering systems, such systems often struggle when additional knowledge is needed to correctly answer the questions. Proposed alternatives involve translating the question and the natural language text to a logical representation and then use logical reasoning. However, this alternative falters  when the size of the text gets bigger. To address this we propose an approach that does logical reasoning over premises written in natural language text. The proposed method uses recent features of Answer Set Programming (ASP) to call external  NLP modules (which may be based on ML) which perform simple textual entailment. To test our approach we develop a corpus based on the life cycle questions and showed that Our system achieves up to $18\%$ performance gain when compared to standard MCQ solvers. 

\end{abstract}

\noindent 
Developing intelligent agents that can understand natural language, reason and use commonsense knowledge has been one of the long term goals of AI. To track the progress towards this goal, several question answering challenges have been proposed \cite{levesque2012winograd,clark2018think,richardson2013mctest,rajpurkar2016squad}. Our work here is related to the school level science question answering challenge, \textsc{Aristo} \cite{clark2015elementary,clark2018think}. As shown in \cite{clark2018think} existing IR based and end-to-end machine learning systems work well on a subset of science questions but there exists a significant amount of  questions that appears to be hard for existing solvers. In this work we focus on one particular genre of such questions, namely questions about life cycles (and more generally, sequences), even though they have a small presence in the corpus.


To get a better understanding of the ``life cycle" questions and the ``hard" ones among them consider the questions from Table \ref{tab:frog}. The text in Table \ref{tab:frog}, which describes the life cycle of a frog does not contain all the knowledge that is necessary to answer the questions. In fact, all the questions require some additional knowledge that is not given in the text. Question $1$ requires knowing the definition of ``middle" of a sequence.  Question $2$ requires the knowledge of ``between". Question 3 on other hand requires the knowledge of ``a good indicator". Note that for question $3$, knowing whether an adult frog has lungs or if it is the adult stage where the frog loses its tail is not sufficient to decide if option (A) is the indicator or option (B). In fact an adult frog satisfies both the conditions. An adult frog has lungs and the tail gets absorbed in the adult stage. It is the uniqueness property that decides that option (B) is an indicator for the adult stage. We believe to answer these questions the system requires access to this knowledge. 

\begin{table}[!t]
		
\begin{tabular}{|p{0.45\textwidth}|}
		\hline
		\multicolumn{1}{|c|}{\textbf{Life Cycle of a Frog}}\\
		\textbf{order}: egg $\rightarrow$ tadpole  $\rightarrow$ tadpole with legs  $\rightarrow$ adult\\\\
		\textbf{egg} - Tiny frog eggs are laid in masses in the water by a female frog. The eggs hatch into tadpoles.\\\\
		\textbf{tadpole} - (also called the polliwog) This stage hatches from the egg. The tadpole spends its time swimming in the water, eating and growing. Tadpoles breathe using gills and have a tail.\\\\
		\textbf{tadpole with legs} - In this stage the tadpole sprouts legs (and then arms), has a longer body, and has a more distinct head. It still breathes using gills and has a tail.\\\\
		\textbf{froglet} - In this stage, the almost mature frog breathes with lungs and still has some of its tail.\\\\
		\textbf{adult} - The adult frog breathes with lungs and has no tail (it has been absorbed by the body).\\\\
	    
	    	    \textbf{1}. {\it What is the middle stage in a frog’s life? (A) tadpole with legs (B) froglet  }\vspace{4pt}

	    	     \textbf{2}. {\it What is a stage that comes between tadpole and adult in the life cycle of a frog?\vspace{4pt}
	    	      (A) egg (B) froglet }

	    	     \textbf{3}. {\it What best indicates that a frog has reached the adult stage? (A) When it has lungs (B) When its tail  has been absorbed by the body }
	    	   \\\hline
		\end{tabular}
		\caption{A text for life cycle of a Frog with few questions.}
		\label{tab:frog}
\end{table}
Since this additional knowledge of ``middle", ``between", ``indicator'' (and some related ones which are shown later) is applicable to any sequence in general and is not specific to only life cycles, we aim to provide this knowledge to the question answering system and then plan to train it so that it can recognize the question types. The paradigm of declarative programming provides a natural solution for adding background knowledge. Also the existing semantic parsers perform well on recognizing questions categories. However the existing declarative programming based question answering methods demand the premises (here the life cycle text) to be given in a logical form. For the domain of life cycle question answering this seems a very demanding and impractical requirement due to the wide variety of sentences that can be present in a life cycle text. Also a life cycle text in our dataset contains $25$ lines on average which makes the translation more challenging. 

The question that we then address is, ``can the system utilize the additional knowledge (for e.g. the knowledge of an ``indicator") without requiring the entire text to be given in a formal language?" We show that by using Answer Set Programming and some of its recent features (function symbols) to call external modules that are trained to do simple textual entailment, it is possible do declaratively reasoning over text. We have developed a system following this approach that answers questions from life cycle text by declaratively reasoning about concepts such as ``middle", ``between", ``indicator'' over premises given in natural language text. To evaluate our method a new dataset has been created with the help of Amazon Mechanical Turk. The entire dataset contains $5811$ questions that are created from 41 life cycle texts. A part of this dataset is used for testing. Our system achieved up to $18\%$ performance improvements when compared to standard baselines. 

Our contributions in this work are two-fold: (a) we propose a novel declarative programming method that accepts natural language texts as premises, which as a result extends the range of applications where declarative programming can be applied and also brings down the development time significantly; (b) we create a new dataset of life cycle texts and questions (\url{https://goo.gl/YmNQKp}), which contains annotated logical forms for each question. 



\section{Background}
\subsection{Answer Set Programming} 
An Answer Set Program is a collection of rules of the form,\vspace{-4pt}
\begin{align*}
	L_{0}  \text{ :- }  L_{1},...,L_m, \textbf{not } L_{m+1},..., \textbf{not } L_n.
\end{align*}
where each of the $L_i$'s is a literal in the sense of classical logic. Intuitively, the above rule means that if 
$L_{1},...,L_m$ are true and if $L_{m+1},..., L_n$ can be safely assumed to be false then 
$L_0$ must be true \cite{gelfond1988stable}. The left-hand side of an ASP rule is called the \textit{head} and the right-hand side is called the \textit{body}. The symbol :- (``if'') is dropped if the \textit{body} is empty; such rules are called facts. Throughout this paper, predicates and constants in a rule start with a lower case letter, while variables start with a capital letter. The following ASP program represents question $3$ from Table \ref{tab:frog} with three facts and one rule.
\vspace{-5pt}
\begin{lstlisting}[label=qrep,caption=a sample question representation,frame=tb]
qIndicator(frog,adult).
option(a, has(lungs)).
option(b, hasNo(tail)).
ans(X):- option(X,V), indicator(O,S,V), 
         qIndicator(O,S).
\end{lstlisting}

\noindent
The first \textit{fact} represents that question $3$ is an `indicator' type question and is looking for something which indicates that a \textit{frog} is in the \textit{adult} stage. The later two \textit{facts} roughly describes the two answer choices, namely ``(a) when it has lungs'' and ``(b) when its tail  has been absorbed by the body''. The last rule describes that for an indicator type question, the option number $X$ is a correct answer if the answer choice $V$ is an indicator for the organism $O$ being in stage $S$ i.e. if $indicator(O,S,V)$ is true.\\

\noindent\textbf{Aggregates}
A rule in ASP can contain aggregate functions. An aggregate function takes as input a set. ASP has four built-in aggregates namely \textit{\#count, \#max, \#min, \#sum} which respectively computes the number of elements in a set, the maximum, minimum or the sum of numbers in the set. The follows rule defines the concept of an `indicator' using the \textit{\#count} aggregate.\vspace{-6pt}
\begin{lstlisting}[label=indicator_normal,caption=Defining Indicator of a stage, frame=tb]
indicator(O,Stage,P) :-
   stageFact(O,Stage,P),
   #count {stageFact(O,S1,P)} = 1.
\end{lstlisting}
Here, $stageFact(O,Stage,P)$ captures the attributes $P$ that are true when the organism $O$ is in the stage $S$. The above rule then describes that $P$ is an indicator for $O$ being in stage $S$ if $P$ is true in $S$ and it is only true in $S$ i.e. the total number of stages $S1$ where $Prop$ is true is one.\\

\noindent\textbf{String valued Terms}
The object constants in ASP can take \textit{string} values (written inside quotes `` ''). This is useful while working with text. For example, the options in the question $3$ can also be represented as follows:\vspace{-4pt}
\begin{lstlisting}
option(a, "when it has lungs").
option(b, "when its tail has 
           been absorbed by the body").
\end{lstlisting} 

\noindent\textbf{Function Symbols}
A \textit{function symbol} allows calling an external function  which is defined in a scripting language such as \textit{lua} or \textit{python} \cite{calimeri2008computable}.  An occurrence of a function symbol making an external call is preceded by the `@' symbol. For e.g., $@stageFact(O,S,P)$ denotes a function symbol that calls to an external function named  $stageFact$ which takes three arguments as input. A function symbol can return any simple term such as \textit{name}, \textit{number} and \textit{strings} as output.

\subsection{QA using Declarative Programming}
A question answering (QA) system that follows declarative programming approach primarily requires three components: a \textbf{semantic parser} $\mathcal{SP}$, a  \textbf{knowledge base} $\mathcal{KB}$ and a \textbf{set of rules} (let's call it \textbf{theory}) $\mathcal{T}$.

\begin{itemize}
\item The goal of the \textbf{semantic parser} $\mathcal{SP}$ is to translate a given question into a logical form. 
\item The $\mathcal{KB}$  provides facts  or ``premises" with respect to which the question should be answered.  For e.g. for the frog life cycle the $\mathcal{KB}$ might look like the following:\vspace{-4pt}
\begin{lstlisting}[frame=tb, label=kb, caption=A sample KB for part of the Frog life cycle]
stageFact(frog,tadpole,has(tail)).
stageFact(frog,froglet,has(lungs)).
stageFact(frog,froglet,has(tail)).
stageFact(frog,adult,has(lungs)).
stageFact(frog,adult,hasNo(tail)).
\end{lstlisting}
\item The \textit{theory} $\mathcal{T}$ contains inference enabling rules. 
\end{itemize}
To answer a question, the system first translates the question into a logical form and then combines that with the $\mathcal{KB}$ 
and the $\mathcal{T}$ to create a consolidated program. The output (models) of which provides the answer. 

For the running example of the `indicator' question ($Q3$ from Table \ref{tab:frog}) if the theory $\mathcal{T}$ contains the rule in listing \ref{indicator_normal}, some semantic parser provides the question representation in listing \ref{qrep}  and the $\mathcal{KB}$  contains the facts in listing \ref{kb}, then the output will contain the deduced fact $ans(b)$ describing that option (B) is the correct answer. This is because, the rule in listing \ref{indicator_normal} will deduce from the $\mathcal{KB}$ that $indicator(frog,adult,hasNo(tail))$ is true. The last rule in listing \ref{qrep} will then conclude that $ans(b)$ is true. Since there is no reason to believe that $indicator(frog,adult,has(lungs))$ is true, $ans(a)$ will not be part of the output (model). The semantics of ASP is based on the stable model semantics \cite{gelfond1988stable}. For further details interested readers can refer to \cite{gebser2012answer,gelfond2014knowledge}. 

\section{Proposed Approach}
The issue in the running example is that it is difficult to get the facts in terms of $stageFact/3$ predicate and in the actual $\mathcal{KB}$ we do not have facts in this format. Rather we have the life cycle texts (Table \ref{tab:frog}) describing the facts. To deal with this we replace such predicates with two external function symbols, namely \textit{generate} and \textit{validate}. 
\begin{description}
\item [Generate] A \textit{generate function} for a predicate takes the arguments of the predicate and returns a textual description of the predicate instance following some template. For example, a generate function for \textit{stageFact} can take \textit{(frog, adult, hasNo(tail))} as input and returns a string such as ``an adult frog has no tail" or if it is for a predicate named \textit{parent}  it can take \textit{(x, y)} and return ``$x$ is a parent of $y$".  

\item [Validate] A \textit{validate function} takes a string describing a proposition (e.g. ``an adult frog has no tail") and validates the truthfulness of the proposition against a $\mathcal{KB}$ containing text (e.g. Table \ref{tab:frog}).  For now let us assume a \textit{validate function} returns $1$ or $0$ depending on whether the proposition is true or false according to the text in the $\mathcal{KB}$. 

\end{description}

\noindent With this transformation the ``indicator" rule from listing \ref{indicator_normal} will look as follows:

\begin{lstlisting}[frame=tb]
indicator(O,Stage,Prop) :- 
  P = @g_StageFact(O,Stage,Prop), 
  @v_StageFact(P) ==1,
  #count { S1: v_StageFact(P1)==1, 
    P1 = @g_StageFact(O,S1,Prop)} == 1. 
\end{lstlisting}
The above rule could be read as follows: $Prop$ denotes that $O$ is in stage $S$ if the natural language description of $StageFact(O,Stage,Prop)$ which is obtained by calling the $g\_StageFact$ function is true according to the $v\_StageFact$ function and also the number of stages $S1$ where the natural language description of $StageFact(O,S1,Prop)$ true is equal to $1$. 

The pair of \textbf{generate-validate} function symbols delegates the responsibility of verifying if a proposition is true or not to an external function and we believe that if the proposition is simple enough and close to the texts described in a $KB$, a simple \textbf{validate} function might be able to compute the truth value with good accuracy. However, one important issue with this rule is that it is not ``safe". In simple words the above rule does not specify what values the variables $O,Stage,Prop,S1$ can take and as a result what to pass as arguments to the $g\_StageFact$ functions is undefined. To mitigate this issue one needs to add some domain predicates which describes the possible values of the unbounded variables. For our question answering task, we have used the predicates that represent the question as domain predicates. The resulting rule, then, looks as follows:

\begin{lstlisting}[frame=tb]
indicator(O,Stage,Prop) :-
  (*\bfseries qIndicator(O,Stage),qOption(X,Prop)*),
  P = @g_StageFact(O,Stage,Prop), 
  @v_StageFact(P) ==1,
  #count { S1: (*\bfseries isAStageOf(S1,O)*),
    v_StageFact(P1)==1, 
    P1 = @g_StageFact(O,S1,Prop)} == 1. 
\end{lstlisting}
\noindent The $isAStageOf(S1,O)$ describes the stages in the life cycle of the organism $O$ and is extracted from the ``order'' field in life cycle texts ( ``Order'' in Table \ref{tab:frog}).

\vspace{-5pt}
\paragraph{On the choices of a Validate function} The task of deciding if a proposition is true based on a given text is a much studied problem in the field of NLP and is known as \textit{textual entailment}. There exist several textual entailment functions. All of which can be used as validate function. However, the textual entailment functions returns a real value between $0$ to $1$ denoting the probability that the proposition is true and thus one needs to decide a threshold value to obtain a boolean validate function. In the implementation of our system we have not used a boolean validate function but used the entailment score as it is. We describe how to use a fuzzy validate function in the next section after describing the life cycle dataset and the representation of the texts.

\section{The Dataset and The Implemented System}
The life cycle question answering dataset contains a total of $41$ texts and $5.8k$ questions. Each text contains a \textit{sequence} which describes the order of stages and a \textit{natural language description} of the life cycle as shown in Table \ref{tab:frog}. The life cycle texts are collected from the internet. The sequence of the stages are manually added by either looking at the associated image in the website that describes the order of the stages or from the headings of the text (Table \ref{tab:frog}).\\

\noindent\textbf{Representing Life Cycle Texts}\\
Each life cycle text is represented in terms of two predicates, namely, \textit{stageAt(URL, O, P, S)} and \textit{description(URL, O, T)}. The $stageAt$ predicate describes that according to the source \textit{URL} (from which the text is collected) the stage that comes at position $P$ in the life cycle of $O$ is $S$. The \textit{description} stores the text that describes the life cycle. The following ASP program shows the representation of the text in Table \ref{tab:frog}. To save space the actual value of the \textit{URL} is replaced by `u'. The $\mathcal{KB}$ contains representations of $41$ such texts.
\begin{lstlisting}
stageAt(u,"frog",1,"egg").
stageAt(u,"frog",2,"tadpole").
stageAt(u,"frog",3,"tadpole with legs")
stageAt(u,"frog",4,"froglet").
stageAt(u,"frog",5,"adult").
description(u,"frog", 
    "Egg: Tiny frog eggs are laid...").
\end{lstlisting}

\subsection{Question Categories}
The question that are created from these texts are divided into $11$ categories. The first three types of questions namely \textit{look up, difference, indicator} require reading the textual description of stages whereas the remaining six types of questions can be answered solely from the sequence of stages (egg $\rightarrow$ tadpole  $\rightarrow$ tadpole with legs  $\rightarrow$ adult).\\

\noindent\textbf{Look Up Questions} This category contains questions the answer to which can be directly looked up from the description of the stages and does not require any special thinking. The following list shows some questions in this category:

\begin{displayquote}
\textit{How do froglets breath?	(A) using lungs	(B) using gills}
\end{displayquote}
\begin{displayquote}
\textit{The tail of a frog disappears at what stage? (A) adult	(B) froglet}
\end{displayquote}
\begin{displayquote}
\textit{Where do female frogs lay their eggs? (A) In water	(B) On land}
\end{displayquote}

\noindent\textbf{Difference Questions}
This category of questions compare two stages based on their physical attributes, abilities or need that is true in one stage but not in other. The following list shows examples:

\begin{displayquote}
\textit{What is an adult newt able to do that a tadpole cannot? (A) walk on land	(B) swim in water}
\end{displayquote}
\begin{displayquote}
\textit{A tadpole just turned into an eft. What does it need now? (A) shade	(B)water}
\end{displayquote}
\begin{displayquote}
\textit{A seedling develops what that a sprout does not have? (A) protective bark (B) root}
\end{displayquote}

\noindent\textbf{Indicator Questions}
This category of questions mentions an organism, a stage, two answer choices and asks which one of those indicates that the organism is in the given stage. Question $3$ in Table \ref{tab:frog} provides an example of this.\\

\noindent\textbf{Sequence Based Questions}
Questions from this category can be answered based on the sequence of stages that describes journey of an organism from beginning to the end (e.g. egg $\rightarrow$ tadpole  $\rightarrow$ tadpole with legs  $\rightarrow$ adult). Questions in this category are further divided into $8$ classes which takes one of the following forms: \begin{enumerate*}[label=(\arabic*)]
\item \textit{Next Stage Questions}: given a stage and an organism, asks for the next stage.
\item \textit{Before Stage Questions}: given a stage and an organism, asks for the stages that appear before.
\item \textit{Between Stage Questions}: given two stages and an organism, asks for the stages that appear between those two.
\item \textit{Stage At Questions}: given an organism and a position, asks for the stages that appear at that position.
\item \textit{Count Questions}: given an organism asks how many stages are there in the life cycle.
\item \textit{Correctly Ordered Questions}: given an organism asks the sequence that describes the correct order of the stages.
\item \textit{Stage Of Questions}: given an organism asks for the stages that appear in its life cycle.
\item \textit{Not a Stage of Questions}: given an organism asks for the stages that do not appear in its life cycle.
\end{enumerate*} Table \ref{tab:examples} shows an example of each types of questions.\\

\begin{table*}[!htb]
	\centering
	\begin{tabular}{ |l|p{135pt}|l|l| }
		\hline
        Question  Template  & Example Question & Instantiated Template & \#Qs \\\hline
        $qLookup(O)$ & How do froglets breath? & $qLookup(``frog")$ &2525\\\hline
        $qDifference(O,S1,S2)$ & What is an adult newt able to do that a tadpole cannot? & $qDifference(``newt",``tadpole",``adult")$ & 167\\\hline
        $qIndicator(O,S)$& When do you consider a penguin to have reached the adult stage?
         & $qIndicator(``penguin",``adult")$ & 125\\\hline
        $qNextStage(O,S)$&A salmon spends time as which of these after emerging from an egg?
                 & $qNextStage(``salmon",``egg")$ & 346\\\hline
        $qStageBefore(O,S)$&Newt has grown enough  but it is  not yet in the tadpole stage, where it might be?&$qStageBefore(``newt",``tadpole")$&123\\\hline
        $qStageBetween(O,S1,S2)$&What is the  stage that comes after egg and before eft in the newt life cycle?
        &$qStageBetween(``newt",``egg",``eft")$&123\\\hline
        $qStageAt(O,P)$ & What stage a longleaf pine will be in when it is halfway through its life?&$qStageAt(``longleaf ~pine",middle)$&520\\\hline
        $ qCorrectlyOrdered(O)$& To grow into an adult, fleas go through several stages. Which of these is ordered correctly?
                &$qCorrectlyOrdered(``flea")$&43\\\hline
        $qCountStages(O)$ & From start to finish, the growth process of a wolf consists of how many steps?
        &$qCountStages(``wolf'')$&113\\\hline
        $qIsAStageOf(O)$ & The growth process of lizards includes which of these?
        &$qIsAStageOf(``lizard")$&1500\\\hline
        $qIsNotAStageOf(O)$ & To grow into an adult, fleas go through 4 stages. Which of these is not one of them?
                        &$qIsNotAStageOf(``flea")$&227\\\hline
        
	\end{tabular}
	\caption{Question templates and total number of questions for each question category. Variables starting with $O,P,S$ respectively refers to an organism, a position and a stage. A position could be a natural number or any member of \{middle, last\}.} 
	\label{tab:examples}

\end{table*}

\noindent\textbf{Question Representation}\\
The representation of a question comprises of four ASP facts. Given an MCQ question of the form ``$\langle$Q?$\rangle$ (A) $\langle$answer choice 1$\rangle$ (B) $\langle$answer choice 2$\rangle$ '', the first three facts are computed trivially as follows:

\begin{lstlisting}
question(``Q?'').
option(a,``answer choice 1'').
option(b,``answer choice 2'').
\end{lstlisting} 
The fourth fact captures the type of the question (i.e. look up, difference etc.) and some associated attributes (i.e. organism, stages, position). For each one of the $11$ types of questions in the dataset there is a fixed template which describes the associated attributes for each type of question. The fourth fact is an instantiation of that template which is computed by a semantic parser. Table \ref{tab:examples} describes the questions templates and shows an example instantiation.

\subsection{Theory}
The \textit{theory} contains a total of $36$ ASP rules, $3$ \textit{generate} functions one for each of the \textit{look up, difference} and \textit{stage indicator} question type and a single \textit{validate} function. The \textit{validate} function, $@validate(Text, Hypothesis)$ takes as input a life cycle \textit{text} and a \textit{hypothesis} (string) and returns a score between $0$ to $1$. The score is computed using a textual entailment function as follows:

$score = max \{ ~textual\_entailment(S, Hypothesis): \\ ~~~~~~~~~~~~~~~~~~~~~~~~~~~~~~~~~S\textit{ is a sentence in } Text\}$
 
\noindent To find the answer, a confidence score $V\in [0,1]$ is computed for each answer option $X$ (denoted by $confidence(X,V)$). The rules in the \textit{theory} computes these confidence scores. The correct answer is the option that gets maximum score. The following rule describes this:

\begin{lstlisting}
ans(X):- option(X,V), confidence(X,V), 
      V == #max {V1:confidence(X1,V1)}.
\end{lstlisting}
Due to limited space we only describe the rules that call entailment functions through function symbols.\\

\noindent\textbf{Lookup Questions}
Given the representation of a lookup question such as: \{\textit{qLookup(``frog'').  question(``How do froglets breathe?''). 
option(a,``using gills''). option(b,``using lungs'').}\}, the following rule computes the confidence score for each option.

\begin{lstlisting}
confidence(X,V):-  
  question(Q), qOption(X,C), 
  H = @generate_lookup(Q,C), 
  qLookup(Org), description(URL,Org,P),
  V = @validate(P,H), 
\end{lstlisting}
While creating the confidence for option ``a'' this rule will call the \textit{generate\_lookup(Q,C)} function with $Q$ = ``How do froglets breathe?'' and $C$ = ``using gills''. The \textit{generate\_lookup} function then returns a hypothesis ``froglets breathe using gills''. The \textit{validate} function then takes the description of the frog life cycle and the hypothesis and verifies if any of the sentence in the text supports the hypothesis: ``froglets breathe using gills''. The confidence score of option ``a'' is the score returned by the \textit{validate} function. Similarly it will compute the confidence score for option ``b''.

The work of \cite{khot2018scitail} presents a function that creates a hypothesis from a question and an answer choice which was used to solve MCQ questions. The \textit{generate\_lookup} function here reuses their implementation.\\

\noindent\textbf{Difference Questions}
Given a difference question (e.g. ``What is an adult newt able to do that a tadpole cannot?'' and an answer choice (e.g. ``walk on land'') a \textit{generate} function returns two hypothesis $H_1$ and $H_2$. (``adult newt able to walk on land'', ``a tadpole cannot walk on land''). The fuzzy truth value for each each hypothesis is computed with the $validate$ function. The product of which is assigned to be the confidence score of the answer choice. A rule is written in ASP to describe the same.\\

\noindent\textbf{Indicator Questions}
When dealing with a fuzzy \textit{validate} function the definition of an \textit{indicator} is modified as follows: Let $v$ be the score for an answer choice $c$ that indicates that the organism $O$ is in stage $S$. If $O$ goes through $n$ stages, $S$ represents the $j$-th stage and $p_i$ is the truth value that $c$ is true in stage $i$, then $v = p_j * \prod_{k=1,k\neq j}^{n}(1-p_k)$. The following five ASP rules are written to describe the same.
\begin{lstlisting}[frame=tb]
stageIndicatorIndex(ID):- 
   stageAt(URL, O, ID, S), 
   qStageIndicator(O,S).

trueForStage(Idx,X,V):-qIndicator(O,S),
  option(X,C),stageAt(URL,O, Idx, S1),
  H = @generate_indicator(S1,C)
  description(URL, O, Text),
  V = @validate(Text, H). 
                  
result(1, X , @product("1.0",V,1,ID)):-    
  trueForStage(O, 1,X,V), 
  stageIndicatorIndex(SRC,ID).
                
result(O, N, X, @product(V1,V2,N,ID)):-    
   result(O, N-1, X , V1), 
   trueForStage(O, N,X,V2),
   stageIndicatorIndex(ID). 
                
confidence(X,V):- res(N, X , V),  
  N = #max {P:stageAt(URL,O, P, S )}. 
\end{lstlisting}
The first rule finds out the index of the stage specified in the question. The second rule computes the truth value $p_i$ (\textit{trueForStage(Idx, X, V)}) for each stage index \textit{Idx} and each option $X$. The last three rules compute the confidence score $v = p_j * \prod_{k=1,k\neq j}^{n}(1-p_k)$ iteratively. Here \textit{product(V1, V2, N, ID)} function returns either  $V1*V2$ or $V1*(1-V2)$ depending on whether $N$ is equal to $ID$. The \textit{generate\_indicator} function follows a simple template. It takes as input a stage such as ``froglet'' and an answer choice, for e.g. ``when it has lungs'' and returns ``In the $\langle${froglet}$\rangle$ stage, $\langle${it has lungs}$\rangle$''. 

\section{Dataset Creation}
We  crowdsourced  the  dataset  of  $5811$ multiple-choice  life cycle  questions with their logical forms with the help of Amazon Mechanical Turk. The workers did not create the logical forms. We collected them using reverse-engineering without  exposing  the workers  to the  underlying  formalism. 

To obtain the sequence based questions we followed the crowdsourcing technique in \cite{wang2015building}. Using $stageAt$ predicates in the $\mathcal{KB}$ and the rules in the \textit{theory} we first computed a database of sequence based facts such as $nextSatge(frog, egg, tadpole)$. We then used a simple grammar to create an MCQ question out of it, for e.g, ``What stage comes after egg stage in frog's life? (A) tadpole (B) adult''. Finally we asked the workers to rephrase these questions as much as possible. Since the seed questions were generated using logical facts we could also compute the logical form and the correct answer beforehand.

To collect indicator type questions we gave the workers a life cycle text and described what is meant by an stage indicator question. Each worker were then asked to create two multiple choice stage indicator questions and write down the correct option and associated stage for each question.  There were two workers working on each text. As a result we got $41\times2\times2 = 164$ questions. We manually removed the questions that did not meet the requirements and finally ended up with $125$ questions. Using the stage name that was written down for each question we were able to compute the logical form $qStageIndicator(organism, stage)$. Similarly, a separate task was created to collect \textit{stage difference} questions where the workers apart from the question and the answer choices also wrote down the two stages that are being compared. Using that we computed the logical form.

To obtain look up questions we  gave the workers a life cycle text and asked them to create free form MCQ questions, which gave us $2710$ questions.  We then manually filtered the questions that should belong to the other $10$ categories and ended up with $2525$ look up questions. Since the question template of a look up question only contains the organism name we did not need any extra supervision to create the logical form.

\section{Related Work}

Many question answering systems \cite{sharma2015towards,mitra2016addressing,mitra2015learning,wang2017logic,lierler2017action,clark2018happened,moldovan2003cogex} have been developed that use declarative programming paradigm. Among these the closest to our work are the works of \cite{lierler2017action,mitra2016addressing,clark2018happened} which try to answer a question with respect to a given text. But to do so they convert the associated text into some action language with existing natural language parsers \cite{bos2008wide,he2017deep,flanigan2014discriminative}. Having a formal representation of the text is helpful but the ability to provide special domain knowledge should not be impaired by the absence of a formal representation of the text. Our work can be considered as a step towards that direction. 
  
Our work is also related to \cite{eiter2006effective,havur2014geometric}. \citeauthor{eiter2006effective} have used function symbols (referred to as \textit{external atoms}) to interface ASP with an ontology language (e.g. OWL) that has different formats and semantics. In \cite{havur2014geometric} function symbols are used to delegate some low level feasibility checks (such as ``is it possible to move left without colliding'') in a robotics application.

The task of textual entailment \cite{dagan2006pascal} and semantic parsing \cite{zelle1996learning} play a crucial role in our work. With access to new datasets both the task have received significant attention \cite{bowman2015large,parikh2016decomposable,chen2018neural,wang2015building,krishnamurthy2017neural}.

Finally, recently there has been a surge of new question answering datasets. Depending on their restrictions on the possible answers they can be divided into three categories: (1) the answer is an exact substring of the text (2) the answer can take values from a fixed which is decided by the training dataset and (3) multiple choice questions. We have used the accuracy of existing science MCQ solvers \cite{khot2018scitail} as baselines in our experiment.

\section{Experiments }
\paragraph{Setup}
To evaluate our system we divide the $41$ texts and the $5811$ questions in two different ways:

\textbf{Text Split :} In this case, we follow the machine comprehension style question answering and divide the $41$ life cycle texts into three sets. The training set then contains $29$ texts and $4k$ associated questions, the dev set contains $4$ texts and $487$ questions and the test set contains $8$ texts with $1368$ questions. Given a text and a MCQ question the task is to find the correct answer choice.

\textbf{Question Split :} In this split we  mimic the open book exam setting and divide the $5.8k$ questions randomly into train, dev and test set each containing $4011$, $579$ and $1221$ questions respectively. Here the knowledge base contains all the texts. Given a MCQ question the task is to find out the correct answer choice with respect to the knowledge base.

\paragraph{Our System}
We experiment with four different textual entailment functions. One of those is a neural network based model \cite{parikh2016decomposable}. The remaining three are variations of n-grams and lexical similarity based model \cite{jijkoun2006recognizing}. The first variation (\textsc{NGram-LS-1}) uses WordNet based lexical similarity. The second variation uses (\textsc{NGram-LS-2}) weighted words \cite{jijkoun2006recognizing} along with simple synonym based similarity. The third variation (\textsc{NGram-LS-3}) uses both word weights and WordNet based lexical similarity.

The semantic parser in \cite{krishnamurthy2017neural}  is trained to obtain the question template instances (e.g. $qIndicator(``frog",``adult")$). We observed that the semantic parser predicts the question types (e.g. $qIndicator$) with high accuracy but often make errors in identifying the associated attributes (e.g. ``adult''). For example it predicts that a given question is of \textit{qStageAt} type with $100\%$ accuracy but fails to identify the associated stage index attribute $38\%$ times. Since the question templates in our dataset is quite simple and only contains one organism name, maximally two stage names or one stage index, we employ a simple search to extract the attributes. The resulting semantic parser then works as follows: it first obtains the question type from the trained parser of \cite{krishnamurthy2017neural}. Then it calls a function with a list containing all the organism names and the question. The function then returns the specified organism based on the first organism name that appears in the question. Similarly it makes subsequent calls for extracting stage names and positions. From now on we refer to the semantic parser in \cite{krishnamurthy2017neural} as ``KDG'' and the customized version as ``Customized-KDG''.\\

\noindent\textbf{Baselines}
We use the performance of the entailment functions as baseline scores. For each option a hypothesis is created by combining the question and the answer choice using the code from \cite{khot2018scitail}, which is then passed to an entailment function to compute the confidence score. A second set of baseline is computed using BiDaF \cite{seo2016bidirectional} which performed well across several machine comprehension tasks. Given a passage and a question, BiDaF returns a substring of the passage as an answer. We then use that substring to compute the confidence score for each option. Two versions of BiDaF is used: BiDaF-1 which is trained on \cite{rajpurkar2016squad} and BiDaF-2 which is trained on both \cite{rajpurkar2016squad,clark2018think}. To make the comparison fair, we have added a sentence of the type ``The $i$-th stage is S" for each $stageAt(O,I,S)$  fact in the $\mathcal{KB}$. Also during the evaluation of ``Question Split" only the necessary life cycle text is given as the passage.   

\begin{table}[!htb]
	\centering
	\begin{tabular}{|l@{}|p{40pt}@{}|p{35pt}@{}| }
		\hline
        System  & Acc(\%) Question Split  & Acc(\%) Text Split\\\hline
        Gold + \cite{parikh2016decomposable} &73.63&78.87\\\hline
        Gold + \textsc{NGram-LS-1} & 78.95&\textbf{84.06}\\\hline
        Gold + \textsc{NGram-LS-2} & 79.20&83.77\\\hline
        Gold + \textsc{NGram-LS-3} & \textbf{79.28}&83.77\\\hline\hline
        KDG + \cite{parikh2016decomposable} &70.60&72.51\\\hline 
        KDG + \textsc{NGram-LS-1} &73.87&\textbf{76.17}\\\hline
        KDG + \textsc{NGram-LS-3}&74.28&75.88\\\hline
        KDG + \textsc{NGram-LS-3} &\textbf{74.61}&76.02\\\hline\hline
        Custom-KDG + \cite{parikh2016decomposable} &72.40&76.68\\\hline
        Custom-KDG + \textsc{NGram-LS-1}&77.07&\textbf{80.70}\\\hline
        Custom-KDG + \textsc{NGram-LS-2}&77.72&80.41\\\hline
        Custom-KDG + \textsc{NGram-LS-3}&\textbf{77.80}&80.48\\\hline\hline
        \cite{parikh2016decomposable}& $53.07$&$51.02$ \\\hline
        \textsc{NGram-LS-1}& $61.29$& $61.25$ \\\hline
        \textsc{NGram-LS-2}& $60.44$ & $58.04$ \\\hline
        \textsc{NGram-LS-3}& \textbf{62.40} & \textbf{61.98} \\\hline
        BidaF-1 & 60.03&57.27\\\hline
        BidaF-2 &58.44&60.20\\\hline
	\end{tabular}
	\caption{The first $12$ rows show the performance of our method with different parsers and entailment functions. The last $6$ rows show the performance of the baseline methods.}
	\label{tab:results}
\end{table}

\noindent\textbf{Results}
Table \ref{tab:results} presents the performance of all the systems on both splits. The first four rows show the accuracy of our system when gold representation of the question is used. This shows the best performance that the system can achieve with the entailment functions at hand; which is $79.28\%$ with the \textsc{NGram-LS-3} entailment function on the ``Question Split" and $84.06\%$ with the \textsc{NGram-LS-2} entailment function on the ``Text Split". The next four rows show the performance with the KDG parser.  The errors made by the parser result in an accuracy drop of $\sim5\%$ on ``Question Split" and a drop of $\sim8\%$ on ``Text Split". However, when the customized-KDG parser is used the accuracy on both the split increases. The best accuracy on ``Text Spit" is $77.8\%$ which is within $1.5\%$ of the achievable best with the entailments at hand. The accuracy drop on ``Text split" also reduces from $\sim8\%$ to $\sim3.3\%$. Among the baseline methods which are shown in the last $6$ rows, the best score is achieved by the \textsc{NGram-LS-3} entailment function which is $15.4\%$ less than the best performance achieved by our system on ``Question Split" and $18.72\%$ less on ``Text Split".

\section{Conclusion}
Developing methods that allow machines to reason with background knowledge with premises written in natural language enhances the  applicability of logical reasoning methods and significantly reduces the effort required in building a knowledge based question answering system. In this work we have presented a method towards this direction by using ASP with textual entailment functions. Experiments show the success of our method. However there is still scope for further improvements with the best accuracy being $80.7\%$. The life cycle dataset and the associated code is publicly available to track the progress towards this direction.    

\paragraph{Acknowledgment} The authors would like to thank Tushar Khot for his tremendous support during the experiments. The research is partially funded by NSF 1816039 and AI2 Key Scientific Challenges Program.
\bibliographystyle{aaai}
\bibliography{fb}

\begin{thebibliography}{}

\bibitem[\protect\citeauthoryear{Bos}{2008}]{bos2008wide}
Bos, J.
\newblock 2008.
\newblock Wide-coverage semantic analysis with boxer.
\newblock In {\em Proceedings of the 2008 Conference on Semantics in Text
  Processing}.
\newblock Association for Computational Linguistics.

\bibitem[\protect\citeauthoryear{Bowman \bgroup et al\mbox.\egroup
  }{2015}]{bowman2015large}
Bowman, S.~R.; Angeli, G.; Potts, C.; and Manning, C.~D.
\newblock 2015.
\newblock A large annotated corpus for learning natural language inference.
\newblock {\em arXiv preprint arXiv:1508.05326}.

\bibitem[\protect\citeauthoryear{Calimeri \bgroup et al\mbox.\egroup
  }{2008}]{calimeri2008computable}
Calimeri, F.; Cozza, S.; Ianni, G.; and Leone, N.
\newblock 2008.
\newblock Computable functions in asp: Theory and implementation.
\newblock In {\em International Conference on Logic Programming}.
\newblock Springer.

\bibitem[\protect\citeauthoryear{Chen \bgroup et al\mbox.\egroup
  }{2018}]{chen2018neural}
Chen, Q.; Zhu, X.; Ling, Z.-H.; Inkpen, D.; and Wei, S.
\newblock 2018.
\newblock Neural natural language inference models enhanced with external
  knowledge.
\newblock In {\em Proceedings of the 56th Annual Meeting of the Association for
  Computational Linguistics (Volume 1: Long Papers)}, volume~1,  2406--2417.

\bibitem[\protect\citeauthoryear{Clark \bgroup et al\mbox.\egroup
  }{2018}]{clark2018think}
Clark, P.; Cowhey, I.; Etzioni, O.; Khot, T.; Sabharwal, A.; Schoenick, C.; and
  Tafjord, O.
\newblock 2018.
\newblock Think you have solved question answering? try arc, the ai2 reasoning
  challenge.
\newblock {\em arXiv preprint arXiv:1803.05457}.

\bibitem[\protect\citeauthoryear{Clark, Dalvi, and
  Tandon}{2018}]{clark2018happened}
Clark, P.; Dalvi, B.; and Tandon, N.
\newblock 2018.
\newblock What happened? leveraging verbnet to predict the effects of actions
  in procedural text.
\newblock {\em arXiv preprint arXiv:1804.05435}.

\bibitem[\protect\citeauthoryear{Clark}{2015}]{clark2015elementary}
Clark, P.
\newblock 2015.
\newblock Elementary school science and math tests as a driver for ai: Take the
  aristo challenge!
\newblock In {\em AAAI}.

\bibitem[\protect\citeauthoryear{Dagan, Glickman, and
  Magnini}{2006}]{dagan2006pascal}
Dagan, I.; Glickman, O.; and Magnini, B.
\newblock 2006.
\newblock The pascal recognising textual entailment challenge.
\newblock In {\em Machine learning challenges. evaluating predictive
  uncertainty, visual object classification, and recognising tectual
  entailment}. Springer.
\newblock  177--190.

\bibitem[\protect\citeauthoryear{Eiter \bgroup et al\mbox.\egroup
  }{2006}]{eiter2006effective}
Eiter, T.; Ianni, G.; Schindlauer, R.; and Tompits, H.
\newblock 2006.
\newblock Effective integration of declarative rules with external evaluations
  for semantic-web reasoning.
\newblock In {\em European Semantic Web Conference},  273--287.
\newblock Springer.

\bibitem[\protect\citeauthoryear{Flanigan \bgroup et al\mbox.\egroup
  }{2014}]{flanigan2014discriminative}
Flanigan, J.; Thomson, S.; Carbonell, J.; Dyer, C.; and Smith, N.~A.
\newblock 2014.
\newblock A discriminative graph-based parser for the abstract meaning
  representation.

\bibitem[\protect\citeauthoryear{Gebser \bgroup et al\mbox.\egroup
  }{2012}]{gebser2012answer}
Gebser, M.; Kaminski, R.; Kaufmann, B.; and Schaub, T.
\newblock 2012.
\newblock Answer set solving in practice.
\newblock {\em Synthesis Lectures on Artificial Intelligence and Machine
  Learning} 6(3):1--238.

\bibitem[\protect\citeauthoryear{Gelfond and Kahl}{2014}]{gelfond2014knowledge}
Gelfond, M., and Kahl, Y.
\newblock 2014.
\newblock {\em Knowledge representation, reasoning, and the design of
  intelligent agents: The answer-set programming approach}.
\newblock Cambridge University Press.

\bibitem[\protect\citeauthoryear{Gelfond and
  Lifschitz}{1988}]{gelfond1988stable}
Gelfond, M., and Lifschitz, V.
\newblock 1988.
\newblock The stable model semantics for logic programming.
\newblock In {\em ICLP/SLP}, volume~88.

\bibitem[\protect\citeauthoryear{Havur \bgroup et al\mbox.\egroup
  }{2014}]{havur2014geometric}
Havur, G.; Ozbilgin, G.; Erdem, E.; and Patoglu, V.
\newblock 2014.
\newblock Geometric rearrangement of multiple movable objects on cluttered
  surfaces: A hybrid reasoning approach.
\newblock In {\em Robotics and Automation (ICRA)},  445--452.
\newblock IEEE.

\bibitem[\protect\citeauthoryear{He \bgroup et al\mbox.\egroup
  }{2017}]{he2017deep}
He, L.; Lee, K.; Lewis, M.; and Zettlemoyer, L.
\newblock 2017.
\newblock Deep semantic role labeling: What works and what’s next.
\newblock In {\em Proceedings of the 55th Annual Meeting of the Association for
  Computational Linguistics (Volume 1: Long Papers)}.

\bibitem[\protect\citeauthoryear{Jijkoun and
  De~Rijke}{2006}]{jijkoun2006recognizing}
Jijkoun, V., and De~Rijke, M.
\newblock 2006.
\newblock Recognizing textual entailment: Is word similarity enough?
\newblock In {\em Machine Learning Challenges. Evaluating Predictive
  Uncertainty, Visual Object Classification, and Recognising Tectual
  Entailment}.

\bibitem[\protect\citeauthoryear{Khot, Sabharwal, and
  Clark}{2018}]{khot2018scitail}
Khot, T.; Sabharwal, A.; and Clark, P.
\newblock 2018.
\newblock Scitail: A textual entailment dataset from science question
  answering.
\newblock In {\em Proceedings of AAAI}.

\bibitem[\protect\citeauthoryear{Krishnamurthy, Dasigi, and
  Gardner}{2017}]{krishnamurthy2017neural}
Krishnamurthy, J.; Dasigi, P.; and Gardner, M.
\newblock 2017.
\newblock Neural semantic parsing with type constraints for semi-structured
  tables.
\newblock In {\em Proceedings of the 2017 Conference on Empirical Methods in
  Natural Language Processing},  1516--1526.

\bibitem[\protect\citeauthoryear{Levesque, Davis, and
  Morgenstern}{2012}]{levesque2012winograd}
Levesque, H.~J.; Davis, E.; and Morgenstern, L.
\newblock 2012.
\newblock The winograd schema challenge.
\newblock In {\em KR}.

\bibitem[\protect\citeauthoryear{Lierler, Inclezan, and
  Gelfond}{2017}]{lierler2017action}
Lierler, Y.; Inclezan, D.; and Gelfond, M.
\newblock 2017.
\newblock Action languages and question answering.
\newblock In {\em IWCS 2017—12th International Conference on Computational
  Semantics}.

\bibitem[\protect\citeauthoryear{Mitra and Baral}{2015}]{mitra2015learning}
Mitra, A., and Baral, C.
\newblock 2015.
\newblock Learning to automatically solve logic grid puzzles.
\newblock In {\em EMNLP},  1023--1033.

\bibitem[\protect\citeauthoryear{Mitra and Baral}{2016}]{mitra2016addressing}
Mitra, A., and Baral, C.
\newblock 2016.
\newblock Addressing a question answering challenge by combining statistical
  methods with inductive rule learning and reasoning.
\newblock In {\em AAAI},  2779--2785.

\bibitem[\protect\citeauthoryear{Moldovan \bgroup et al\mbox.\egroup
  }{2003}]{moldovan2003cogex}
Moldovan, D.; Clark, C.; Harabagiu, S.; and Maiorano, S.
\newblock 2003.
\newblock Cogex: A logic prover for question answering.
\newblock In {\em Proceedings of the 2003 Conference of the North American
  Chapter of the Association for Computational Linguistics on Human Language
  Technology-Volume 1}.

\bibitem[\protect\citeauthoryear{Parikh \bgroup et al\mbox.\egroup
  }{2016}]{parikh2016decomposable}
Parikh, A.~P.; T{\"a}ckstr{\"o}m, O.; Das, D.; and Uszkoreit, J.
\newblock 2016.
\newblock A decomposable attention model for natural language inference.
\newblock {\em arXiv preprint arXiv:1606.01933}.

\bibitem[\protect\citeauthoryear{Rajpurkar \bgroup et al\mbox.\egroup
  }{2016}]{rajpurkar2016squad}
Rajpurkar, P.; Zhang, J.; Lopyrev, K.; and Liang, P.
\newblock 2016.
\newblock Squad: 100,000+ questions for machine comprehension of text.
\newblock {\em arXiv preprint arXiv:1606.05250}.

\bibitem[\protect\citeauthoryear{Richardson, Burges, and
  Renshaw}{2013}]{richardson2013mctest}
Richardson, M.; Burges, C.~J.; and Renshaw, E.
\newblock 2013.
\newblock Mctest: A challenge dataset for the open-domain machine comprehension
  of text.
\newblock In {\em EMNLP}, volume~1, ~2.

\bibitem[\protect\citeauthoryear{Seo \bgroup et al\mbox.\egroup
  }{2016}]{seo2016bidirectional}
Seo, M.; Kembhavi, A.; Farhadi, A.; and Hajishirzi, H.
\newblock 2016.
\newblock Bidirectional attention flow for machine comprehension.
\newblock {\em arXiv preprint arXiv:1611.01603}.

\bibitem[\protect\citeauthoryear{Sharma \bgroup et al\mbox.\egroup
  }{2015}]{sharma2015towards}
Sharma, A.; Vo, N.~H.; Aditya, S.; and Baral, C.
\newblock 2015.
\newblock Towards addressing the winograd schema challenge-building and using a
  semantic parser and a knowledge hunting module.
\newblock In {\em IJCAI}.

\bibitem[\protect\citeauthoryear{Wang, Berant, and
  Liang}{2015}]{wang2015building}
Wang, Y.; Berant, J.; and Liang, P.
\newblock 2015.
\newblock Building a semantic parser overnight.
\newblock In {\em Proceedings of the 53rd Annual Meeting of the Association for
  Computational Linguistics and the 7th International Joint Conference on
  Natural Language Processing (Volume 1: Long Papers)}, volume~1.

\bibitem[\protect\citeauthoryear{Wang, Lee, and Kim}{2017}]{wang2017logic}
Wang, Y.; Lee, J.; and Kim, D.~S.
\newblock 2017.
\newblock A logic based approach to answering questions about alternatives in
  diy domains.
\newblock In {\em AAAI},  4753--4759.

\bibitem[\protect\citeauthoryear{Zelle and Mooney}{1996}]{zelle1996learning}
Zelle, J.~M., and Mooney, R.~J.
\newblock 1996.
\newblock Learning to parse database queries using inductive logic programming.
\newblock In {\em Proceedings of the national conference on artificial
  intelligence}.

\end{thebibliography}
\end{document}